# Quantifying Inherent Randomness in Machine Learning Algorithms

Soham Raste, Rahul Singh, Joel Vaughan, and Vijayan N. Nair
Corporate Model Risk, Wells Fargo

## Abstract

Most machine learning (ML) algorithms have several stochastic elements, and their performances are affected by these sources of randomness. This paper uses an empirical study to systematically examine the effects of two sources: randomness in model training and randomness in the partitioning of a dataset into training and test subsets. We quantify and compare the magnitude of the variation in predictive performance for the following ML algorithms: Random Forests (RFs), Gradient Boosting Machines (GBMs), and Feedforward Neural Networks (FFNNs). Among the different algorithms, randomness in model training causes larger variation for FFNNs compared to tree-based methods. This is to be expected as FFNNs have more stochastic elements that are part of their model initialization and training. We also found that random splitting of datasets leads to higher variation compared to the inherent randomness from model training. The variation from data splitting can be a major issue if the original dataset has considerable heterogeneity.

There are several implications for practitioners:
- When performance results of multiple ML algorithms are compared, one must keep in mind that part of any observed differences can be due to inherent randomness.
- Comparisons of ML algorithms in the literature or with deterministic algorithms typically do not control or account for such variation. So users should be cautious with claims that one algorithm is better than another due to small differences in performances.
- It is recommended that users provide some measure of variation from the source of randomness, together with the performance metrics, to allow a proper comparison of algorithms.
- In ML, the random seed used for model initialization and training is an integral part of the algorithm, and it should be reported along with the results so that they can be reproduced.
- The variation from randomness in data splitting should be controlled by using the same training and test subsets for the different algorithms that are being compared. This is also true if one uses cross-validation to assess performance.

## 1  Introduction

The increased application of machine learning (ML) algorithms has led to more scrutiny of their properties, such as reproducibility, stability, robustness, and interpretability. The results of an algorithm are said to be reproducible if different users fitting the same algorithm to the same set of data get (essentially) the same fitted results and performances. Most of the traditional statistical algorithms are deterministic in nature, so their results are reproducible. However, common ML algorithms have stochastic elements that induce variation in the outputs (predicted model and hence performance measures). For example, a key element of Random Forests (RFs) involves random bootstrap sampling of the rows to create multiple datasets. This leads to different data subsets and hence variation in the results. We will discuss the sources of randomness for the different algorithms in Section 2.

A second major source of variation arises from random partitioning of the dataset into training and test subsets. Traditionally, the assessment of statistical models was done using metrics, such as $R^2$, confidence intervals, $p$ −values, and so on, all computed on the entire dataset. In ML, the performance of ML algorithms is assessed by: i) splitting the dataset (typically randomly) into training and test subsets, ii) developing the model on training data, and iii) assessing performance on the held-out test dataset (see Figure 1). Thus, the performance results will depend on the particular test dataset selected.

There is a third source of variation that arises from hyper-parameter optimization: the random choice of validation dataset and any stochastic elements inherent in the optimization process. We discuss this only briefly in the present paper and leave it for future study.

Figure 1 shows in green the two sources of variation that we focus on: i) random initialization of model training; and ii) random partition of a dataset into training and test subsets. We consider three supervised ML algorithms, RF, GBM, and FFNN, as well as several of their variations, and systematically study the effects of randomness on model results and quantify the magnitude of variation.

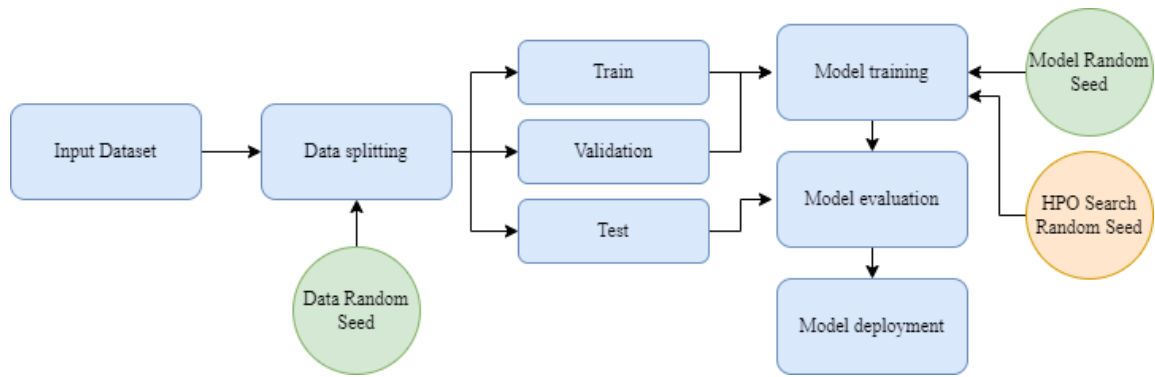

**Figure 1: ML model building process**

There has been some investigation of the impact of randomness on deep neural networks (DNNs). Paper [1] studied the effect of random initialization and examined prediction and interpretation stability. Paper [2] quantified the disagreements between predictions as 'churn' and proposed methods to mitigate the problem of churn in deep neural networks.

The rest of this paper is organized as follows. Section 2 discusses the sources of variation controlled by random seeds in selected machine learning models. The next section describes the experimental setup and methodology. Section 4 describes our empirical results based on several datasets and ML algorithms. Section 5 briefly discusses the impact of random seeds on the selection of optimal hyper-parameters. The paper concludes with our findings in Section 6.

## 2  Background

### 2.1  Variation from Model Training of Algorithms

We consider three commonly used supervised ML algorithms for tabular data: FFNN; RF; GBM, and a popular variation of GBM called XGB. Readers are referred to the literature for details on how these algorithms are trained. See, for example, [3] for a brief overview and additional references. All these ML algorithms have inherent randomness in their training process. Typically, this is controlled by a **random seed** that is set at the start of model training. Once this is fixed, the random number generator produces same sequence of pseudo random numbers, and hence the training algorithm can deliver reproducible results, provided other environmental factors are all the same. In the rest of this sub-section, we discuss the different parts of an algorithm that are controlled by the random seed.

**a) FFNN**

There are several stochastic elements associated with neural networks.
  1) Random initialization of weights: FFNNs have weights associated with the neurons in each layer, and the algorithm learns the optimal weights using gradient descent. However, the weights have to be initialized, and this is done randomly using methods like *He initialization* [3] or *Xavier initialization* [4].
  2) Data shuffling prior to each training epoch: Mini-batch gradient descent is commonly used, where the parameters are updated after each batch of inputs is passed through the network. The training dataset is randomly shuffled prior to each training epoch to remove the effect of data order on the output.
  3) Early stopping: This is a common practice to prevent overfitting. The stopping criterion is based on the validation score calculated on a randomly sampled validation set after every epoch.
  4) Dropout [5]: This is another method of preventing overfitting. During training, outputs from some number of layers are randomly set to zero. This is somewhat like training a large number of neural networks with different architectures in parallel. The number of neurons to be dropped is determined by dropout ratio parameter.

**b) RF**

There are two sources of randomness with RFs, and both are controlled by a single random seed.
  1) Row subsampling**:** Given a training dataset, multiple training sets are created by randomly subsampling rows with replacement (bootstrap sampling).
  2) Column subsampling: To decrease the correlation between trees grown on the different subsets, random column subsampling is also typically performed. More precisely, only the subset of the features in the selected columns are used in the tree.

**c) GBM**

The sources of randomness in GBM are also column and row subsampling, as with RF. But some implementations of GBM do not use either or both.

**d) XGBoost**

Extreme gradient boosting (XGBoost [6]) is a variant of GBM that uses more regularized forms of gradient boosting with L1 and L2 regularizations. We study it here as a separate method, as it is perhaps the most commonly used implementation of GBM. Like GBM, XGBoost can also have <u>row and column subsampling</u> which will induce randomness.

As we noted earlier, the above stochastic elements for each algorithm are controlled by the selection of a single random seed before training the model. We will call this as <u>model seed</u> from now on. There are potentially other sources of variation during model training. For example, if the model is trained on a GPU, the results have other stochastic elements, such as those from how the parallel processing is done. We do not consider these aspects of variation in our study, and all our models were trained on CPUs. In addition, we controlled other aspects of the comparisons. For FFNN, we used version '2.4.1' of tensorflow keras [7] framework, and for XGB model we used version '1.2.0' of xgboost package [8]. For GBM and RF algorithms, we used version '0.24.1' of scikit-learn framework [9].

### 2.2 Variation from Random Partitioning of Dataset

As shown in Figure 1, another key step in ML model development is to split the original dataset into training, validation, and test subsets. The training and validation sets are used during model training and hyper-parameter tuning respectively, while the test set is used to assess performance. The typical practice is to randomly split the original set according to some predefined proportions (such as 0.6; 0.2; 0.2). This random data splitting is also controlled by a random seed that we will refer to <u>data seed</u> in this paper. If this seed is fixed, we obtain exactly the same subsets for training, validation, and test. We will study the effect of this source of variation by systematically varying the data seed with the model seed fixed.

### 2.3 Variation from Hyper-parameter Tuning

ML algorithms have hyper-parameters that have to be tuned (optimized) during model training. There are several strategies for hyper-parameter tuning [10]. The simplest is grid search, where the performance is evaluated for all possible combinations of hyper-parameters from pre-defined search space. Grid search tends to be wasteful [16], and random search of the hyper-parameter space [11] is typically preferred. In this approach, a random subset of all possible combinations is searched to arrive at the optimal configuration of hyper-parameters. This induces another source of randomness since the final optimal hyper-parameters obtained will depend on which random subset was selected. We discuss this briefly in Section 5.

## 3 Experimental Setup

### 3.1 Algorithms Considered

| Algorithms | Description |
|---|---|
| FFNN | All standard practices are considered. Dropout and shuffling prior to training epochs are allowed. |
| FFNN-Res | Restricted FFNN with no dropout or shuffling |
| RF | Both row and column subsampling |
| GBM | Both row and column subsampling |
| XGB | Both row and column subsampling |

**Table 1: Algorithms**

Table 1 summarizes the different algorithms we studied. FFNN allows all the sources of randomness in Table 1 to vary, whereas FFNN-Res has no dropout or shuffling. Only the vanilla version of RF with both row and column sub-sampling was considered. For GBM and XGB, the versions without row and column subsampling lead to deterministic results (no variation with model seeds) so we did not consider them in our comparisons.

### 3.2 Datasets

We used four different datasets to study the variation in predicted performances: two with binary response and two with continuous response (see Table 2). For all the datasets, we performed min-max scaling of the features to bring all the values into (0, 1) range.

| Name | Source | Response | #samples | #features | Class distribution for binary response |
|---|---|---|---|---|---|
| Home Mortgage | Internal | Binary | 1 million | 55 | Label 1: 1% Label 0: 99% |
| MAGIC Gamma Telescope | UCI repository | Binary | 19,020 | 10 | Label 1: 65% Label 0: 35% |
| Bike Sharing | UCI repository | Continuous | 17,389 | 10 | - |
| California Housing | UCI repository | Continuous | 20,640 | 8 | - |

**Table 2: Description of Datasets**

**a) Home Mortgage**: This dataset deals with defaults of home mortgage loans. The goal is to predict the probability of troubled loans (label 1) based on a several predictors such as credit score, unemployment rate, delinquency status, loan to value ratio (LTV), etc.

**b) MAGIC Gamma Telescope**: This is a publicly available dataset [12]. It is simulated data of registrations of high-energy gamma particles in a gamma telescope. There are 10 numerical features which describe the shower image patterns in the form of attributes of the ellipse (width, length, etc.) and attributes of pixels (size, concentration, etc.). The task is to classify whether the shower image was produced by gamma radiation (label 1) or background noise (label 0).

**c) Bike Sharing**: This dataset contains hourly counts of bike rentals for two years [12]. There was also information on temperature, humidity, wind speed, season, day-of-the-week, working day vs holiday etc. The goal is to predict the number of bike rentals by hour of day. We transformed the response variable (counts) in the bike sharing dataset to log-counts in our analysis.

**d) California Housing:** This is from the 1990 U.S. census [12] where each observation corresponds a census block group, typically 600 to 3,000 residents. The response variable is median house value for the block. The features included longitude, latitude, housing median age, medium income, population, total rooms, total bedrooms and households.

### 3.3 Description of Study

The study was designed as follows (Figure 2). We considered the four datasets in Section 3.2 and the five algorithms in Section 3.1. In addition, we considered four different data split ratios of test subsets: are 0.1, 0.2, 0.3, and 0.4. For every combination of dataset, algorithm, and data split ratio, we did the following:

i) Select a random model seed and a random data seed. Split the dataset randomly based on the data seed, train the algorithm with the model seed, tune the hyper-parameters for the algorithm using five-fold cross-validation, and fix these optimal parameter combinations for the steps below. See Appendix for the details of the hyper-parameter search space for the different algorithms. We used random search with 500 trials to get the best hyper-parameter configuration.

ii) <u>Fix data seed</u> as 0 and select 50 random model seeds. Retrain the model 50 times with 50 different model seeds and with fixed hyper-parameters as obtained in Step i). Evaluate the performance on test set for each of the 50 runs.

iii) <u>Fix model seed</u> as 0 and select 50 random data seeds. Retrain the model 50 times with 50 different data seeds and with fixed hyper-parameters as obtained in Step i). Evaluate the performance on the respective test sets for each of the 50 runs.

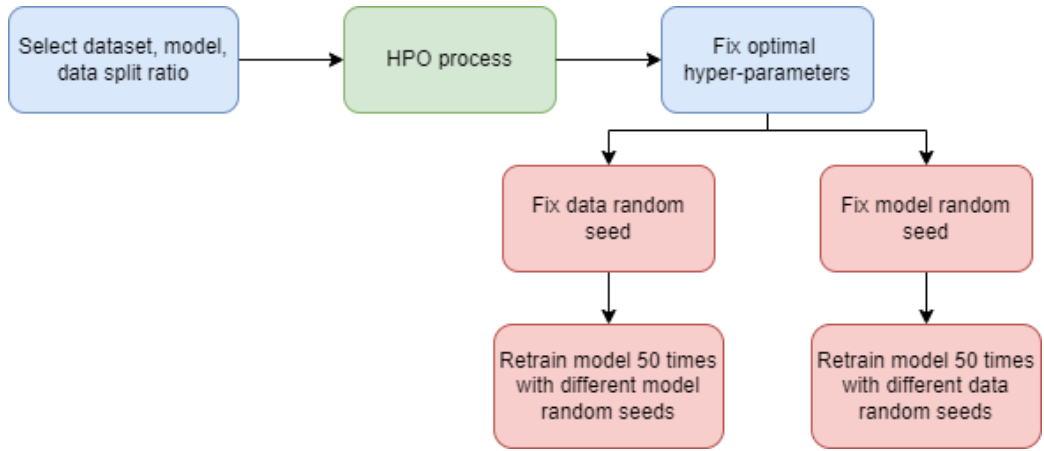

Figure 2: Experimental Setup

## 4 Results and Discussion

### 4.1 Variation from Model and Data Seeds

Before describing the full study, we note the results of an initial experiment for the California housing dataset. We fixed the model seed and data seed and repeated the training and assessment 50 times each. There was no variation in the results, confirming that fixing these random seeds yields reproducible results and any variation that we observed subsequently must be due to randomness in data and model seeds.

For the full study, we conducted experiments with changing model and data seeds separately. We summarize the results for the four different datasets in Table 3. These are for the two different cases: one with changing data seeds and the other with changing model seeds. Here, the data split ratio was fixed at 0.2. The metrics reported in the table are as follows:

1. <u>Median</u>: median test performance (MSE/AUC) out of 50 different runs
2. <u>IQR</u>: Inter-quartile range (upper quartile – lower quartile)
3. <u>Range</u>: maximum − minimum
4. <u>Relative Variation</u>: (Range)/(Median)

| Bike Sharing | | | | | | | | |
|---|---|---|---|---|---|---|---|---|
| Algorithm | Data Seed | | | | Model Seed | | | |
| | Median MSE | IQR | Range | Relative Variation | Median MSE | IQR | Range | Relative Variation |
| FFNN | 0.1218 | 0.0053 | 0.0260 | 0.213 | 0.1246 | 0.0029 | 0.0098 | 0.079 |
| FFNN-Res | 0.1290 | 0.0100 | 0.0324 | 0.251 | 0.1318 | 0.0050 | 0.0254 | 0.184 |
| RF | 0.1277 | 0.0089 | 0.0255 | 0.200 | 0.1331 | 0.001 | 0.0050 | 0.038 |
| GBM | 0.1088 | 0.0064 | 0.0166 | 0.153 | 0.1131 | 0.0013 | 0.0052 | 0.046 |
| XGB | 0.1088 | 0.0059 | 0.0178 | 0.164 | 0.1128 | 0.0013 | 0.0037 | 0.033 |

| California Housing | | | | | | | | |
|---|---|---|---|---|---|---|---|---|
| Algorithm | Data Seed | | | | Model Seed | | | |
| | Median MSE | IQR | Range | Relative Variation | Median MSE | IQR | Range | Relative Variation |
| FFNN | 0.2532 | 0.0158 | 0.0621 | 0.245 | 0.2321 | 0.0050 | 0.0185 | 0.080 |
| FFNN-Res | 0.2711 | 0.0250 | 0.1288 | 0.475 | 0.2607 | 0.0120 | 0.0484 | 0.185 |
| RF | 0.2564 | 0.0155 | 0.0525 | 0.205 | 0.2617 | 0.0017 | 0.0058 | 0.022 |
| GBM | 0.2052 | 0.0100 | 0.0410 | 0.200 | 0.1974 | 0.0020 | 0.0073 | 0.037 |
| XGB | 0.2109 | 0.0122 | 0.0408 | 0.193 | 0.2003 | 0.0030 | 0.0133 | 0.066 |

| MAGIC Gamma Telescope | | | | | | | | |
|---|---|---|---|---|---|---|---|---|
| Algorithm | Data Seed | | | | Model Seed | | | |
| | Median AUC | IQR | Range | Relative Variation | Median AUC | IQR | Range | Relative Variation |
| FFNN | 0.9333 | 0.0062 | 0.0151 | 0.016 | 0.9337 | 0.0014 | 0.0061 | 0.006 |
| FFNN-Res | 0.9282 | 0.0052 | 0.0182 | 0.020 | 0.9281 | 0.0020 | 0.0060 | 0.006 |
| RF | 0.9215 | 0.0060 | 0.0142 | 0.015 | 0.9138 | 0.0004 | 0.0018 | 0.002 |
| GBM | 0.9362 | 0.0058 | 0.0159 | 0.017 | 0.9316 | 0.0011 | 0.0037 | 0.004 |
| XGB | 0.9374 | 0.0049 | 0.0144 | 0.015 | 0.9317 | 0.0014 | 0.0045 | 0.005 |

| Home Mortgage | | | | | | | | |
|---|---|---|---|---|---|---|---|---|
| Algorithm | Data Seed | | | | Model Seed | | | |
| | Median AUC | IQR | Range | Relative Variation | Median AUC | IQR | Range | Relative Variation |
| FFNN | 0.8430 | 0.0049 | 0.0187 | 0.022 | 0.8427 | 0.0027 | 0.0096 | 0.011 |
| FFNN-Res | 0.8411 | 0.0070 | 0.0181 | 0.021 | 0.8432 | 0.0037 | 0.0126 | 0.015 |
| RF | 0.8534 | 0.0047 | 0.0181 | 0.021 | 0.8511 | 0.0003 | 0.0011 | 0.001 |
| GBM | 0.8581 | 0.0046 | 0.0177 | 0.020 | 0.8578 | 0.0009 | 0.0028 | 0.003 |
| XGB | 0.8608 | 0.0047 | 0.0170 | 0.020 | 0.8609 | 0.0010 | 0.0037 | 0.004 |

**Table 3: Summary of Results**

We first discuss the variation due to model seeds observed in Table 3.
- FFNN and FFNN-Res have high variation (significantly higher in most cases) compared to tree ensemble algorithms.
- FFNN-Res, the restricted version without dropout and shuffling, shows more variability than FFNN despite having only weight initialization as the source of randomness. Furthermore, the median performance of FFNN is better than FFNN-Res in most of the cases. So it appears that that dropout and shuffling help FFNN to arrive at a better performance. The results on variation suggest that the performances are also more stable.
- For the tree-based models, RF shows least variability with respect to changing model seeds. This might be because the final prediction is an average or majority vote of a large number of trees, which smooths out the variation. GBM and XGB have comparable levels of variation.
- While we are not focusing on predictive performances in this paper, the columns on median MSE and AUC show that XGB/GBM generally perform the best with FFNN being close. See [13] for a more extensive study of comparative performance of these ML algorithms.

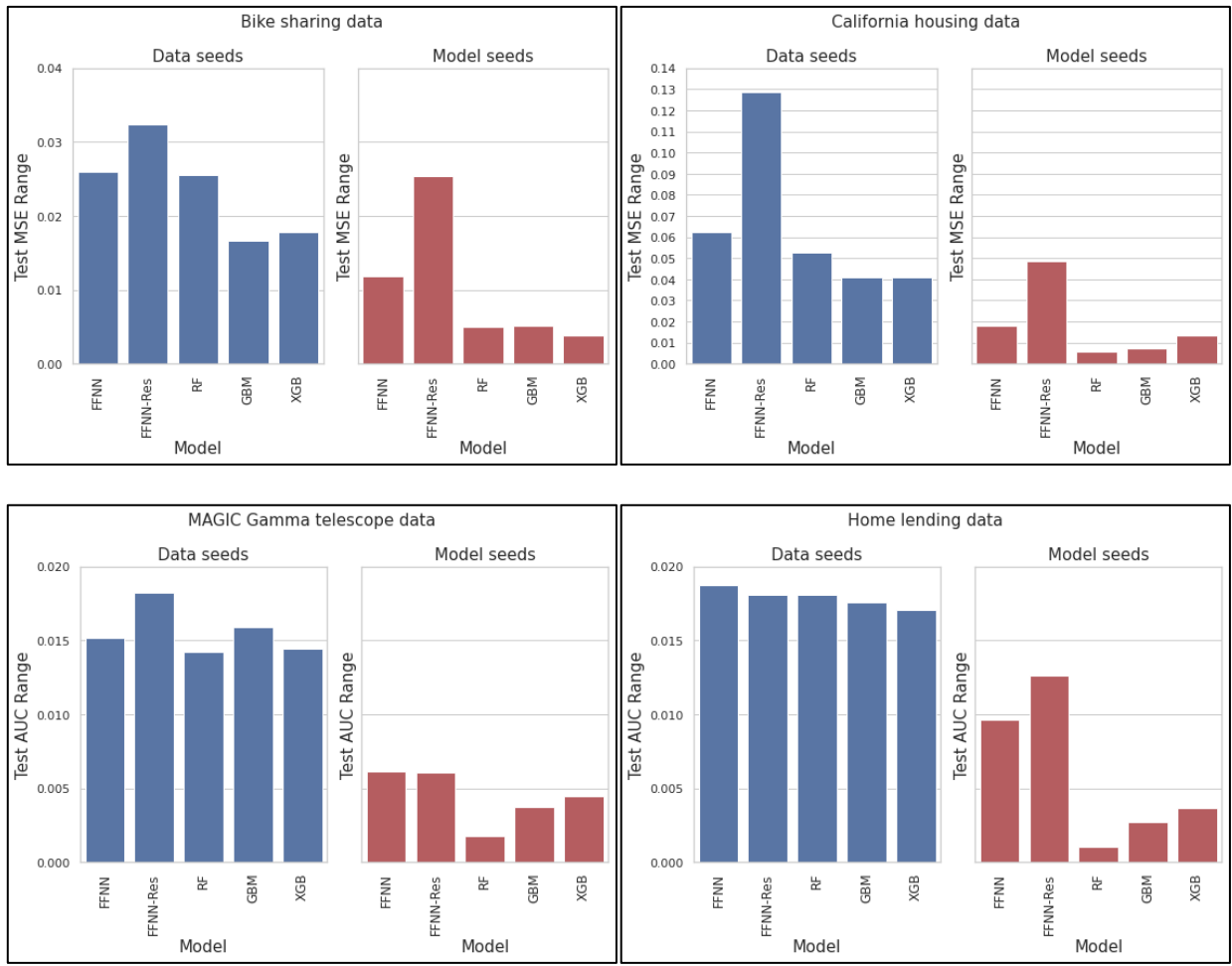

**Figure 3: Range = (Max – Min) of Test MSEs and Test AUCs (over 50 random seeds) for the four different datasets**

Turning now to variation due to data seeds, we observe from Table 3 that:
- The variation from changing data seeds is much higher compared to changing model seeds.
- However, the differences across the algorithms are much smaller. This is particularly true among the tree-based algorithms.

- The results for FFNN are comparable to the tree-based algorithms.
- FFNN-Res has bigger variability in some instances.

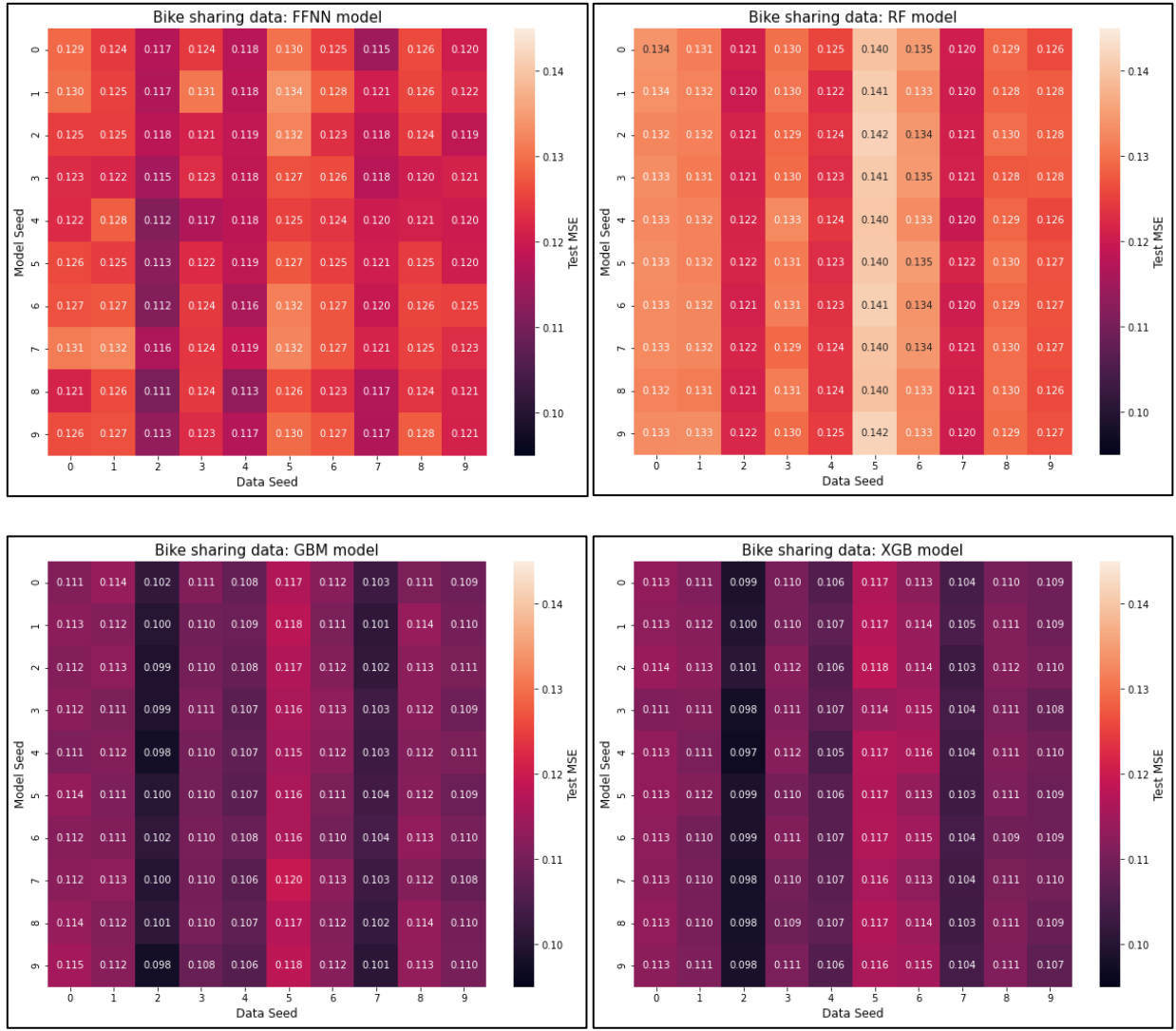

**Figure 4: Heat maps of Test MSEs for Bike Sharing data for the four different ML algorithms**
Each of the $10 \times 10 = 100$ cells represents MSEs trained with different pairs of model and data seeds

As noted above, changes in data seeds cause higher compared to changing model seeds. This is more easily seen from Figure 3, where the bar plots in red correspond to model seeds and those in blue to data seeds. The panels show range (max – min) of the 50 test MSEs or AUCs with either the model seeds or the data seeds changed. It is clear that changing test data subsets) leads to higher variation than that from randomness in model training. This would be of particular concern if there is a lot of heterogeneity in the original dataset.

Figure 4 provides an in-depth look at the difference in variation between data seeds and model seeds for Bike Share data. Each heat map shows the Test MSEs for the $10 \times 10 = 100$ pairs of data and model seeds. We see that the colors do not change much as we vary the model seeds (rows) for a fixed data seed (column). On the other hand, there are distinct differences in colors as we vary the data seeds (columns) with model seed (row) fixed. In fact, there is considerable consistency within columns, implying that changing model seeds within a column (data seed) does not lead to as much variation. This phenomenon seems to be consistent across the algorithms as seen in Figure 4. For example, data seeds 2, 4 and 7 deliver best performance (lowest MSE) for all the algorithms, whereas data seed 5 gives worst performance.

**4.2  Changing Data-Splitting Ratios**

So far, the data split ratio was fixed at 0.2 for the test datasets. In this section, we vary the ratio from 0.1, 0.2., 0.3, and 0.4 to assess the effect of changing data-splitting ratios. The details are as follows:

   i)    For every combination of (dataset, algorithm, and split-ratio), hyper-parameter optimization (HPO) was performed;
   ii)   The obtained hyper-parameters were fixed, and we conducted two experiments;
   iii)  In one experiment, model seeds were changed while the data seed was fixed;
   iv)   In the next experiment, the data seeds were changed and model seed was fixed.
   v)    Both experiments were performed for each of the four data split ratios.

| California Housing: Model seeds varied | | | | | | | | | | | |
|---|---|---|---|---|---|---|---|---|---|---|---|
| Data split ratio | FFNN | | | RF | | | GBM | | | XGB | |
| | Median MSE | Range | | Median MSE | Range | | Median MSE | Range | | Median MSE | Range |
| 0.1 | 0.2277 | 0.0481 | | 0.2568 | 0.0072 | | 0.1893 | 0.0114 | | 0.1874 | 0.0103 |
| 0.2 | 0.2426 | 0.0181 | | 0.2617 | 0.0058 | | 0.1974 | 0.0073 | | 0.2003 | 0.0133 |
| 0.3 | 0.2493 | 0.0190 | | 0.2710 | 0.0051 | | 0.2048 | 0.0058 | | 0.2111 | 0.0095 |
| 0.4 | 0.2518 | 0.0215 | | 0.2733 | 0.0065 | | 0.2168 | 0.0052 | | 0.2170 | 0.0082 |

| California Housing: Data seeds varied | | | | | | | | | | | |
|---|---|---|---|---|---|---|---|---|---|---|---|
| Data split ratio | FFNN | | | RF | | | GBM | | | XGB | |
| | Median MSE | Range | | Median MSE | Range | | Median MSE | Range | | Median MSE | Range |
| 0.1 | 0.2535 | 0.0792 | | 0.2566 | 0.0681 | | 0.2031 | 0.0597 | | 0.2053 | 0.0616 |
| 0.2 | 0.2532 | 0.0521 | | 0.2564 | 0.0525 | | 0.2052 | 0.0410 | | 0.2109 | 0.0408 |
| 0.3 | 0.2451 | 0.0417 | | 0.2619 | 0.0407 | | 0.2096 | 0.0349 | | 0.2110 | 0.0260 |
| 0.4 | 0.2544 | 0.0396 | | 0.2655 | 0.0267 | | 0.2134 | 0.0240 | | 0.2128 | 0.0273 |

**Table 4: Summary of results for different data-split ratios for California Housing Data**

Table 4 shows the results for different data split ratios for the four algorithms. The columns correspond to median MSE over the 50 seeds and the range (= max – min) of the MSEs. We show results for California Housing data only. The results for others were qualitatively similar. The variation (range) from changing model seeds does not change considerably across data-split ratios. This is the case for all algorithms. On the other hand, the variation from changing data seeds decreases as data-split ratio increases. The latter observation is intuitively reasonable because, as the ratio increases, the test subsets will have more overlap. Thus, variation due to changing data seeds will be smaller.

## 5   Variation from randomness in hyper-parameter tuning

The results so far were based on tuning the hyper-parameters once initially and fixing them subsequently in all the experiments (see Figure 2). In practice, however, for each pair of model and random seed, we have to tune the hyper-parameters each time. In this section, we report the results from a limited study to assess this source of randomness. Results from a more comprehensive study will be reported in the future.

We considered only the MAGIC Gamma Telescope dataset. For each pair of $10 \times 10 = 100$ model and data seeds, we tuned the hyper-parameters, obtained the "best" model, and then evaluated the test performance. The search spaces for HPO configurations are given in the Appendix. We used random search strategy with 500 trials. Table 5 shows the results from the median test AUC and variation across the data and random seeds. Unlike Section 4, the results in this section capture the variation in the entire modeling process: variation due to changing hyper-parameters, model seeds, and data seeds. Table 5 shows a comparison of the results from fixed hyper-parameters experiment (Section 4) and tuned hyper-parameters experiment (Section 5). Looking at both the range and relative variation, we do not see substantial difference in variation for the two cases. This is true for both model seed and data seed experiments.

| MAGIC Gamma Telescope: Model seeds varied | | | | | | | | |
|---|---|---|---|---|---|---|---|---|
| Algorithm | Fixed hyper-parameters | | | | Tuned hyper-parameters | | | |
| | Median AUC | IQR | Range | Relative Variation | Median AUC | IQR | Range | Relative Variation |
| FFNN | 0.9337 | 0.0014 | 0.0061 | 0.006 | 0.9302 | 0.0018 | 0.0062 | 0.007 |
| RF | 0.9138 | 0.0004 | 0.0018 | 0.002 | 0.9132 | 0.0006 | 0.0013 | 0.001 |
| GBM | 0.9316 | 0.0011 | 0.0037 | 0.004 | 0.9318 | 0.0022 | 0.0036 | 0.004 |
| XGB | 0.9317 | 0.0014 | 0.0045 | 0.005 | 0.9327 | 0.0021 | 0.0031 | 0.003 |

| MAGIC Gamma Telescope: Data seeds varied | | | | | | | | |
|---|---|---|---|---|---|---|---|---|
| Algorithm | Fixed hyper-parameters | | | | Tuned hyper-parameters | | | |
| | Median AUC | IQR | Range | Relative Variation | Median AUC | IQR | Range | Relative Variation |
| FFNN | 0.9333 | 0.0062 | 0.0151 | 0.016 | 0.9316 | 0.0042 | 0.0105 | 0.011 |
| RF | 0.9215 | 0.0060 | 0.0142 | 0.015 | 0.9213 | 0.0046 | 0.0136 | 0.015 |
| GBM | 0.9362 | 0.0058 | 0.0159 | 0.017 | 0.9346 | 0.0043 | 0.0094 | 0.010 |
| XGB | 0.9374 | 0.0049 | 0.0144 | 0.015 | 0.9368 | 0.0038 | 0.0102 | 0.011 |

**Table 5: Comparison of variation in performances: Fixed hyper-parameters vs tuned hyper-parameters**

## 6   Conclusions

The inherent sources of randomness are often neglected when a ML algorithm is developed or multiple algorithms are compared. This paper used an empirical study to quantify the magnitude of variation in model training and selection of test datasets. Random seeds are integral part of ML algorithms. To produce reproducible results, the exact details of the algorithm, including the random seeds used, should be reported along with other settings. In addition, we recommend that users train ML algorithms with multiple model and data seeds and provide some measure of variation from these sources of randomness for a proper assessment.

# 7 Appendix

**Hyper-parameter search space**

The details of HPO search space for each algorithm are given here. These were typically used in our study except for some slight changes for particular datasets.

**FFNN**

| Hyper-parameter | Tuned or Fixed | Search space |
| --- | --- | --- |
| Learning rate | Tuned | [0.01, 0.001, 0.0001] |
| Neurons in layer 1 | Tuned | [16, 32, 64, 128] |
| Neurons in layer 2 | Tuned | [16, 32, 64, 128] |
| Neurons in layer 3 | Tuned | [0, 8, 16, 32, 64] |
| Dropout ratio | Tuned | [0.1, 0.2, 0.3] |
| L1 regularization | Tuned | [0. 0.00001, 0.0001, 0.001] |
| L2 regularization | Tuned | [0. 0.00001, 0.0001, 0.001, 0.01] |
| Batch size | Tuned | [512, 1024, 2048, 4096] |
| Maximum Epochs | Fixed | 5000 |
| Early stopping patience | Fixed | 100 |
| Shuffling prior to training epoch | Fixed | True |
| Optimizer | Fixed | Adam |

**FFNN_Res**

| Hyper-parameter | Tuned or Fixed | Search space |
| --- | --- | --- |
| Learning rate | Tuned | [0.01, 0.001, 0.0001] |
| Neurons in layer 1 | Tuned | [16, 32, 64, 128] |
| Neurons in layer 2 | Tuned | [16, 32, 64, 128] |
| Neurons in layer 3 | Tuned | [0, 8, 16, 32, 64] |
| Dropout ratio | Fixed | 0 |
| L1 regularization | Tuned | [0. 0.00001, 0.0001, 0.001] |
| L2 regularization | Tuned | [0. 0.00001, 0.0001, 0.001, 0.01] |
| Batch size | Tuned | [512, 1024, 2048, 4096] |
| Maximum Epochs | Fixed | 5000 |
| Early stopping patience | Fixed | 100 |
| Shuffling prior to training epoch | Fixed | False |
| Optimizer | FIxed | Adam |

**RF**

| Hyper-parameter | Tuned or Fixed | Search space |
|---|---|---|
| Max depth | Tuned | 5 to 15 |
| Number of estimators | Tuned | 50 to 500 |
| Min samples leaf | Tuned | 1 to 100 |
| Max samples (row subsampling) | Tuned | [0.5, 0.6, 0.7, 0.8, 0.9] |
| Max features (column subsampling) | Tuned | [0.3, 0.4, 0.5, 0.6, 0.7, 0.8, 0.9] |

**GBM**

| Hyper-parameter | Tuned or Fixed | Search space |
|---|---|---|
| Learning rate | Tuned | [0.05, 0.1, 0.15, 0.2] |
| Max depth | Tuned | 3 to 5 |
| Number of estimators | Tuned | 50 to 500 |
| Min samples leaf | Tuned | 1 to 100 |
| Subsample (row subsampling) | Tuned | [0.5, 0.6, 0.7, 0.8, 0.9] |
| Max features (column subsampling) | Tuned | [0.3, 0.4, 0.5, 0.6, 0.7, 0.8, 0.9] |

**XGB**

| Hyper-parameter | Tuned or Fixed | Search space |
|---|---|---|
| Learning rate | Tuned | [0.05, 0.1, 0.15, 0.2, 0.25, 0.3] |
| Max depth | Tuned | 3 to 5 |
| Number of estimators | Tuned | 50 to 500 |
| lambda | Tuned | [0, 0.001, 0.1, 1, 10, 100] |
| alpha | Tuned | [0, 0.001, 0.1, 1, 10, 100] |
| Subsample (row subsampling) | Tuned | [0.5, 0.6, 0.7, 0.8, 0.9] |
| Colsample_bytree (column subsampling) | Tuned | [0.3, 0.4, 0.5, 0.6, 0.7, 0.8, 0.9] |